\pdfoutput=1

\documentclass[11pt]{article}
\usepackage[normalem]{ulem}

\usepackage{acl}

\usepackage{times}
\usepackage{latexsym}
\usepackage{graphicx}
\usepackage{booktabs}
\usepackage{multirow}
\usepackage{amsmath}
\usepackage[T1]{fontenc}

\usepackage[utf8]{inputenc}

\usepackage{microtype}

\title{CS-UM6P at SemEval-2022 Task 6: Transformer-based Models  for Intended Sarcasm Detection in English and Arabic}

\author{{\bf Abdelkader {El Mahdaouy}} \hspace{0.5cm} {\bf Abdellah {El Mekki}}  \hspace{0.5cm} {\bf Kabil Essefar}$^{\dagger}$\\ \textbf{Abderrahman Skiredj}$^{\dagger}$ \hspace{0.5cm} \textbf{Ismail Berrada}$^{\dagger}$ \\
School of Computer Science, Mohammed VI Polytechnic University, Morocco \\
{\tt abdelkader.elmahdaouy@um6p.ma, abdellah.elmekki@um6p.ma}
\\
{\tt \{firstname.lastname\}@um6p.ma}$^{\dagger}$
}

\begin{document}
\maketitle
\begin{abstract}
Sarcasm is a form of figurative language where the intended meaning of a sentence differs from its literal meaning. This poses a serious challenge to several Natural Language Processing (NLP) applications such as Sentiment Analysis, Opinion Mining, and Author Profiling. In this paper, we present our participating system to the intended sarcasm detection task in English and Arabic languages. Our system\footnote{The source code of our system is available at \url{https://github.com/AbdelkaderMH/iSarcasmEval}} consists of three deep learning-based models leveraging two existing pre-trained language models for Arabic and English. We have participated in all sub-tasks. Our official submissions achieve the best performance on sub-task A for Arabic language and rank second in sub-task B. For sub-task C, our system is ranked 7th and 11th on Arabic and English datasets, respectively.        
\end{abstract}

\section{Introduction}
Sarcasm is an important aspect of human natural language. It is characterized by the occurrence of a discrepancy between the intended and the literal meaning of utterance \citep{WILSON20061722}. The prevalence of this phenomenon may jeopardize the performance of many NLP applications, such as Sentiment Analysis, Opinion Mining, and Emotion Detection, among others \citep{maynard-greenwood-2014-cares, rosenthal-etal-2014-semeval, van-hee-etal-2018-semeval}. Indeed, sarcasm detection has been the subject of many systematic investigation, where several shared tasks have been organized and a number of datasets have been introduced \cite{van-hee-etal-2018-semeval,10.1145/3368567.3368585, ghosh-etal-2020-report, abu-farha-etal-2021-overview}. The existing datasets are either labeled by a human annotator or using distant supervision signals such as the presence of a set of predefined hashtags \citep{oprea-magdy-2020-isarcasm}. However, these labeling methods might be sub-optimal for intended sarcasm detection as the author's sarcastic intent may differ from an annotator's perceived meaning \citep{oprea-magdy-2019-exploring, oprea-magdy-2020-isarcasm}. Besides, most existing research works have focused on English language \citep{van-hee-etal-2018-semeval, ghosh-etal-2020-report}, while few studies have been introduced for other languages such as Arabic \citep{10.1145/3368567.3368585,abu-farha-etal-2021-overview}. 

To overcome the aforementioned limitations, \citet{abufarha-etal-2022-semeval} have organized the iSarcasmEval shared task for intended sarcasm detection in English and Arabic languages. In contrast to previous research work, the introduced dataset, for intended sarcasm detection, is labeled by the authors themselves. The authors are then asked to provide non-sarcastic rephrases that covey the same intended meaning of their sarcastic texts. Further, the iSarcasmEval's organizers have relied on linguistic experts to categorize sarcastic texts into sarcasm, irony, satire, understatement, overstatement, and rhetorical questions \citep{Leggitt2000}. 

In this paper, we present our participating system to iSarcasmEval shared task \citep{abufarha-etal-2022-semeval}. Our system rely on three transformer-based deep learning models. In all our models, we use existing Pre-trained Language Model (PLM) to encode the input text and apply a single attention layer to the contextualized word embedding of PLM \citep{barbieri2021xlmtwitter, abdul-mageed-etal-2021-arbert}. For all our models, we use the same classifier architecture composed of one hidden layer and one classification layer. The classifier is fed with the concatenation of the pooled output of the PLM as well as the attention layer output. We model the sub-task A as a binary classification (first model) and as a multi-class classification (second model). Motivated by the small size of the datasets and similarly to GAN-BERT architecture \citep{croce-etal-2020-gan}, the third model uses a conditional generator that tries to generate fake samples that are similar to the PLM's embedding of the real input data. The discriminator of the third model is trained to discriminate between fake and real samples as well as classify the real ones correctly. For sub-task B, we only employ the GAN based model (third model), while for sub-task C, we utilize all  models trained on Task A and we compare the probabilities of the sarcastic class of the left and the right text. We train our models using several loss functions, including the focal loss \cite{abs-1708-02002}. Besides, for sub-task A, we train our models with and without the non-sarcastic rephrase texts.     

For the official submissions, we employ hard voting ensemble of our trained models. Our system achieve promising results as we rank  1st, 15th, 2nd, 7th, and 11th on sub-task A AR, sub-task A EN, sub-task B, sub-task C AR, and sub-task C EN, respectively. 

\section{Background}

\subsection{Task description}
The organizers of iSarcasmEval shared task have provided training data and testing data for intended sarcasm detection in English and Arabic languages \citep{abufarha-etal-2022-semeval}. The datasets are collected from Twitter and labeled for intended sarcasm detection by the authors of the tweets. For English, the training data consists of 3,468 samples out of which 867 samples are sarcastic. For Arabic, the training data contains 3,102 samples, where 745 samples are sarcastic. The organizers also provide the non-sarcastic rephrase of sarcastic texts for both languages and the dialect of the given Arabic tweets. The shared task consists of the flowing sub-tasks:

\begin{itemize}
    \item \textbf{Sub-task A} is a binary classification task, where the aim is to determine if a tweet is sarcastic or not. This sub-task consists of two sub-tasks A EN and A AR. The test data contains 1,400 for each language. 
    
    \item \textbf{Sub-task B} is a multi-label classification task, where the aim is to assign a given tweet into the sarcasm, irony, satire, understatement, overstatement, and rhetorical question categories of ironic speech \citep{Leggitt2000}. This task is provided for English language only and the test data contains 1,400 samples.
    
    \item \textbf{Sub-task C} aims to 
    identify the sarcastic tweet and the non-sarcastic rephrase given two texts that convey the same meaning. This sub-task consists of two sub-tasks C EN and C AR. The test sets of both languages consist of 200 samples.   
\end{itemize}

\subsection{Related work}

In recent years, there has been a growing number of research works focusing on fine-tuning the existing PLMs on NLP tasks. These PLMs are based on the transformer architecture and are trained using self-supervised learning objectives such as Masked Language Modeling (MLM) amongst others \citep{devlin-etal-2019-bert}.  Several multilingual and monolingual PLM variants are introduced \citep{devlin-etal-2019-bert,  conneau-etal-2020-unsupervised, antoun-etal-2020-arabert}. For domain-specific data, domain adaptive fine-tuning of existing PLMs using MLM or domain adaptation have been shown to improve the performance of NLP applications \citep{rietzler-etal-2020-adapt,barbieri2021xlmtwitter,el-mekki-etal-2021-domain}. Nevertheless, when the domain-specific data is sufficiently large, these transformers can be trained from scratch \citep{abdul-mageed-etal-2021-arbert,inoue-etal-2021-interplay}.

For sarcasm detection, several research studies have been introduced based on fine-tuning the existing PLMs for English and Arabic languages  \citep{10.1145/3368567.3368585, ghosh-etal-2020-report, abu-farha-etal-2021-overview}.  \citet{el-mahdaouy-etal-2021-deep} have shown that incorporating attention layers on top of the contextualized word embedding of the PLM improves the performance of multi-task and single-task learning models for both sarcasm detection and sentiment analysis in Arabic. The main idea consists of classifying the input text based on the concatenation of the PLM's pooled output and the output of the attention layer. This Architecture has yielded promising results on other tasks such as detecting and rating humor, lexical complexity prediction, and fine-grained Arabic dialect identification   \citep{essefar-etal-2021-cs,el-mamoun-etal-2021-cs,el-mekki-etal-2021-bert}.  

Although transformer-based architectures have shown state-of-the-art performance on many NLP tasks, their task-specific fine-tuning requires a reasonable amount of labeled data. Nevertheless, in real-world applications, one may not always have enough labeled training data. Motivated by the performance of Semi-Supervised Generative Adversarial Networks (SS-GAN) in computer vision \citep{odena2016semisupervised}, \citet{croce-etal-2020-gan} have introduced GAN-BERT. The latter extends BERT fine-tuning procedure by training a generator and a discriminator. The generator is trained to generate fake samples embeddings that are similar to the real input text embeddings, whereas, the discriminator is trained to detect fake examples and categorize the real text inputs.

\section{System overview}
The submitted system to the iSarcasmEval shared task relies on three transformer-based models for intended sarcasm detection. In order to encode the input text, we utilize the Twitter-XLM-Roberta-base \citep{barbieri2021xlmtwitter} and MARBERT \citep{abdul-mageed-etal-2021-arbert} for English and Arabic texts, respectively. The former is a variant of XLM-RoBERTa PLM that is adapted to Twitter data using MLM objective, while the latter is a variant of BERT PLM that is pre-trained from scratch on Arabic tweet corpora.  In the following subsection, we describe the components of our system.

\subsection{Preprocessing}
The tweet preprocessing component spaces out emojis and substitutes user's mention and URL with their special tokens of the PLM's tokenizer. For Twitter-XLM-Roberta-base, URLs and user's mentions are replaced by 'http' and '@user'. For MARBERT, they are replaced by 'user' and 'url' special tokens. To leverage the dialect information for Arabic data, we replace the dialect string with its full Arabic name and pass the input text to MARBERT's tokenizer as follows:
\begin{itemize}
    \item [$\bullet$] [SEP] dialect [SEP] preprocessed text [SEP]
\end{itemize}

\subsection{Deep Learning Models}
Our three deep learning models are described as follows:

\begin{itemize}
    \item \textbf{Model 1} consists of a transformer encoder, one attention layer, and a classifier. Following the work of \cite{el-mahdaouy-etal-2021-deep}, we apply attention to the contextualized word embedding of the encoder. The classifier is composed of one hidden layer and one classification layer for binary classification. The classifier is fed with the concatenation of the PLM's pooled output and the attention layer's output.
    
    \item \textbf{Model 2} is similar is to \textbf{Model 1} and the task is modeled as multi-class classification problem. In other words, the classification layer of this model consists of two hidden units. Figure \ref{fig:arch} illustrates the overall architecture of Model 1 and Model 2.
    
    \begin{figure}
        \centering
        \includegraphics[scale=0.95]{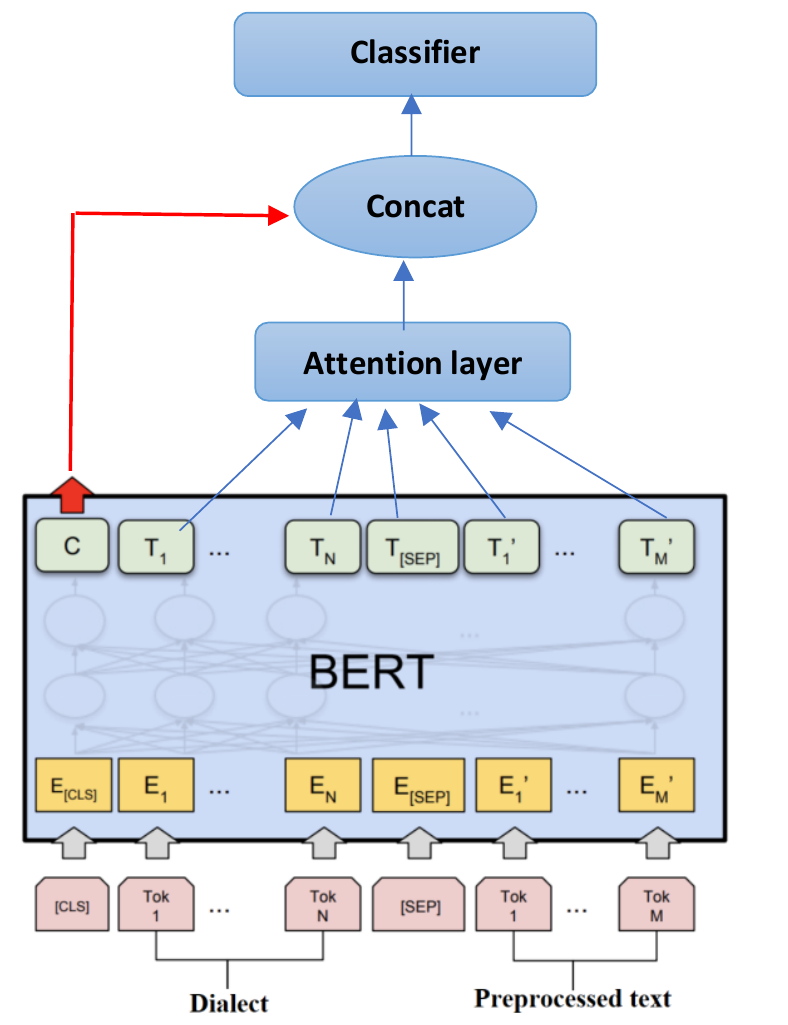}
        \caption{The overall architecture of Model 1 and Model 2.}
        \label{fig:arch}
    \end{figure}
    
    \item \textbf{Model 3} is similar to GAN-BERT model \citep{croce-etal-2020-gan}, whereas, we employ a conditional generator that generates fake embeddings from a random noise and the class category. The model 3 consits of three components, a BERT-based encoder with an extra attention layer on top of the contextualized word embedding, a generator, and a discriminator.  The encoder  represents the input sentences using the CLS token embedding and the output of the attention layer.  The generator is trained to fool the discriminator by generating fake embedding representations that are similar to the input text embedding. It consists of two hidden layers and one output layer. Each hidden layer is flowed by a dropout layer and the relu activation layer. The discriminator is trained to discriminate between fake examples and real ones and to classify real input texts into sarcastic and non-sarcastic labels. The discriminator is also composed of two hidden layers and one classification layer. Similarly to the generator, the hidden layers are followed by one dropout layer and the relu activation layer.   
\end{itemize}
\subsection{Training objectives}
\label{sec:obj}
For training our models, we utilizes several loss functions, including the Focal Loss \citep{abs-1708-02002}. The aim is to assess the performance of the following training objectives under class imbalance:
\begin{itemize}
    \item For \textbf{Model 1}, we investigate the Binary Cross-Entropy loss (BCE), the Weighted Binary Cross-Entropy loss (W. BCE), and the Binary Focal Loss (BFL). For the W. BCE loss, the positive class weight is set to: $$\frac{batch\_size - positive\_count}{positive\_count + \epsilon}$$
    
    \item For \textbf{Model 2 and 3}, we investigate the Cross-Entropy loss, the Weighted Cross-Entropy loss (W. CE), and the Focal Loss (FL). For the W. CE, the positive and the negative class weights are computed as follows:
\begin{equation*}
\centering
\begin{cases}
pos\_weight = \frac{batch\_size - positive\_count}{positive\_count + \epsilon} \\
neg\_weight = \frac{batch\_size - negative\_count}{negative\_count + \epsilon} 
\end{cases}
 \end{equation*}
\end{itemize}

\section{Experimental setup}

\begin{table*}[!htb]
  \centering
  \caption{The obtained results on the test set of sub-task A for both Arabic and English. For our non official submissions, we report the macro F-1 score of the sarcastic class only.}
  \resizebox{1\textwidth}{!}{%
    \begin{tabular}{c|cccccc|cccccc}
\cmidrule{2-13}    \multicolumn{1}{r}{} & \multicolumn{6}{c|}{\textbf{Sub-Task A Arabic}}       & \multicolumn{6}{c}{\textbf{Sub-Task A English}} \\
\cmidrule{2-13}    \multicolumn{1}{r}{} & \multicolumn{3}{c|}{Tweet only} & \multicolumn{3}{c|}{Tweet + rephrase} & \multicolumn{3}{c|}{Tweet only} & \multicolumn{3}{c}{Tweet + rephrase} \\
\cmidrule{2-13}    \multicolumn{1}{r}{} & BCE/CE & BFL/FL    & \multicolumn{1}{c|}{W. BCE/CE} & BCE   & BFL/FL    & W. BCE/CE & BCE/CE & BFL/FL    & \multicolumn{1}{c|}{W. BCE/CE} & BCE/CE & BFL/FL    & W. BCE/CE \\
    \midrule
    \multicolumn{1}{l|}{Model 1} & 0.5253 & \textbf{0.5621} & \multicolumn{1}{c|}{0.4565} & \textbf{0.6135} & 0.5787 & \textbf{0.5793} & 0.3485 & \textbf{0.3605} & \multicolumn{1}{c|}{0.3421} & \textit{\textbf{0.3833}} & 0.3313 & \textbf{0.3714} \\
    \multicolumn{1}{l|}{Model 2} & 0.5307 & 0.4481 & \multicolumn{1}{c|}{0.4892} & 0.5949 & \textit{\textbf{0.6217}} & 0.5505 & \textbf{0.3574} & 0.3375 & \multicolumn{1}{c|}{0.3144} & 0.3770 & \textbf{0.3619} & 0.3090 \\
    \multicolumn{1}{l|}{Model 3} & \textbf{0.56} & 0.5271 & \multicolumn{1}{c|}{\textbf{0.4937}} & 0.5339 & 0.5488 & 0.5345 & 0.3470 & 0.3448 & \multicolumn{1}{c|}{\textbf{0.3478}} & 0.3517 & 0.3517 & 0.3557 \\
    \midrule
    \midrule
    \multicolumn{1}{r}{} & \multicolumn{6}{c|}{\textbf{Official Submission}} & \multicolumn{6}{c}{\textbf{Official Submission}} \\
    \midrule
    \multirow{2}[4]{*}{\textbf{Ensembling}} & F-1 sarcastic  & F-score  & Precision  & Recall  & Accuracy  & \textbf{Rank} & F-1 sarcastic  & F-score  & Precision  & Recall  & Accuracy  & \textbf{Rank} \\
\cmidrule{2-13}          & 0.5632 & 0.7188 & 0.6948 & 0.8362 & 0.8050 & \textbf{1} & 0.3713 & 0.6171 & 0.6058 & 0.6458 & 0.7750 & \textbf{15} \\
    \bottomrule
    \end{tabular}%
    }%
  \label{tab:taskA}%
\end{table*}%

\begin{table*}[!htb]
  \centering
  \caption{The obtained results on the test set of sub-task B.}
  \resizebox{1\textwidth}{!}{%
    \begin{tabular}{c|c|ccccccc}
\cmidrule{3-9}    \multicolumn{1}{r}{} & \multicolumn{1}{r}{} & \multicolumn{1}{l}{F-1 Macro} & \multicolumn{1}{l}{F-1 sarcasm} & \multicolumn{1}{l}{F-1 irony} & \multicolumn{1}{l}{F-1 satire} & \multicolumn{1}{l}{F-1 understatement} & \multicolumn{1}{l}{F-1 overstatement} & \multicolumn{1}{l}{F-1 rhetorical  question} \\
    \midrule
    \multirow{3}[2]{*}{Model 3} & \multicolumn{1}{l|}{BCE} & \textbf{0.0924} & \textbf{0.2331} & 0.1676 & 0.0530 & 0     & 0     & \textbf{0.1008} \\
          & \multicolumn{1}{l|}{BFL} & 0.0877 & 0.2298 & \textbf{0.1733} & 0.025 & 0     & 0     & 0.0983 \\
          & \multicolumn{1}{l|}{W. BCE} & 0.0681 & 0.2302 & 0.0705 & \textbf{0.0581} & 0     & 0     & 0.0501 \\
    \midrule
    \midrule
    \multirow{2}[4]{*}{\textbf{Ensembling}} & \textbf{Rank} & \multicolumn{7}{c}{\textbf{Official Submission}} \\
\cmidrule{2-9}          & \textbf{2} & 0.0875 & 0.2314 & 0.1622 & 0.0392 & 0     & 0     & 0.0923 \\
    \bottomrule
    \end{tabular}%
    }%
  \label{tab:taskb}%
\end{table*}%

\begin{table*}[!htb]
  \centering
  \caption{The obtained results on the test set of sub-task C for both Arabic and English. For our non-official submissions, we report the accuracy score only.}
  \resizebox{1\textwidth}{!}{%
    \begin{tabular}{c|llcrrr|llcrrr}
\cmidrule{2-13}    \multicolumn{1}{r}{} & \multicolumn{6}{c|}{\textbf{Sub-Task C Arabic}}       & \multicolumn{6}{c}{\textbf{Sub-Task C English}} \\
\cmidrule{2-13}    \multicolumn{1}{r}{} & \multicolumn{3}{c|}{Tweet only} & \multicolumn{3}{c|}{Tweet + rephrase} & \multicolumn{3}{c|}{Tweet only} & \multicolumn{3}{c}{Tweet + rephrase} \\
\cmidrule{2-13}          & \multicolumn{1}{c}{BCE/CE} & \multicolumn{1}{c}{BFL/FL} & \multicolumn{1}{c|}{W. BCE/CE} & \multicolumn{1}{c}{BCE} & \multicolumn{1}{c}{FL} & \multicolumn{1}{c|}{W. BCE/CE} & \multicolumn{1}{c}{BCE/CE} & \multicolumn{1}{c}{BFL/FL} & \multicolumn{1}{c|}{W. BCE/CE} & \multicolumn{1}{c}{BCE/CE} & \multicolumn{1}{c}{BFL/FL} & \multicolumn{1}{c}{W. BCE/CE} \\
    \midrule
    \multicolumn{1}{l|}{Model 1} & \multicolumn{1}{c}{\textbf{0.68}} & \multicolumn{1}{c}{\textbf{0.69}} & \multicolumn{1}{c|}{0.61} & \multicolumn{1}{c}{0.815} & \multicolumn{1}{c}{\textbf{0.82}} & \multicolumn{1}{c|}{0.83} & \multicolumn{1}{c}{\textbf{0.69}} & \multicolumn{1}{c}{\textbf{0.68}} & \multicolumn{1}{c|}{\textbf{0.67}} & \multicolumn{1}{c}{0.695} & \multicolumn{1}{c}{\textbf{0.7}} & \multicolumn{1}{c}{\textbf{0.685}} \\
    \multicolumn{1}{l|}{Model 2} & \multicolumn{1}{c}{0.65} & \multicolumn{1}{c}{0.575} & \multicolumn{1}{c|}{0.59} & \multicolumn{1}{c}{\textbf{0.835}} & \multicolumn{1}{c}{0.815} & \multicolumn{1}{c|}{\textit{\textbf{0.85}}} & \multicolumn{1}{c}{0.655} & \multicolumn{1}{c}{\textbf{0.68}} & \multicolumn{1}{c|}{0.65} & \multicolumn{1}{c}{0.685} & \multicolumn{1}{c}{0.67} & \multicolumn{1}{c}{0.655} \\
    \multicolumn{1}{l|}{Model 3} & \multicolumn{1}{c}{0.67} & \multicolumn{1}{c}{0.68} & \multicolumn{1}{c|}{\textbf{0.685}} & \multicolumn{1}{c}{\textbf{0.835}} & \multicolumn{1}{c}{0.815} & \multicolumn{1}{c|}{0.835} & \multicolumn{1}{c}{0.655} & \multicolumn{1}{c}{\textbf{0.68}} & \multicolumn{1}{c|}{0.625} & \multicolumn{1}{c}{\textit{\textbf{0.71}}} & \multicolumn{1}{c}{0.685} & \multicolumn{1}{c}{\textbf{0.685}} \\
    \midrule
    \midrule
          & \multicolumn{6}{c|}{\textbf{Official Submission}} & \multicolumn{6}{c}{\textbf{Official Submission}} \\
    \midrule
    \multirow{2}[4]{*}{\textbf{Ensembling}} & Accuracy & F-1 Score & \textbf{Rank} &       &       &       & Accuracy & F-1 Score & \textbf{Rank} &       &       &  \\
\cmidrule{2-13}          & 0.7800 & 0.7688  & \textbf{7} &       &       &       & 0.6950 & 0.6481 & \textbf{11} &       &       &  \\
    \bottomrule
    \end{tabular}%
    }
  \label{tab:taskc}%
\end{table*}%

All our models are implemented using the PyTorch\footnote{\url{https://pytorch.org/}} framework and the open-source Transformers\footnote{\url{https://huggingface.co/transformers/}} libraries. Experiments are conducted on a PowerEdge R740 Server having 44 cores Intel Xeon Gold 6152 2.1GHz, a RAM of 384 GB, and a single Nvidia Tesla V100 with 16GB of RAM. $20\%$ of the training set is used for the model validation. All our models are trained using Adam optimizer with a linear learning rate scheduler. Based on our preliminary results, obtained on the validation set, the learning rate, the number of epochs, and the batch size are fixed to $1 \times 10^{-5}$, $10$, and $16$ respectively. For the focal loss, the hyper-parameters $\gamma$ and $\alpha$ (the weight of the negative class) are set to $2$ and $0.8$ respectively. All models are evaluated using the Accuracy as well as the macro averaged Precision, Recall, and F1 measures. Besides, we train our models with and without rephrase texts for sub-task A. For sub-task B, our models are trained on sarcastic tweets.

\section{Results}

In this section, we present the obtained results of our models as well as our official submissions. It is worth mentioning that for sub-task C, we employ the trained models on sub-task A and we use their output probabilities to discriminate between the sarcastic text and the non-sarcastic rephrase. Besides, for our official submissions, we use the hard vote ensemble of our trained models. For each loss function, the best performance obtained is highlighted in \textbf{bold} font, while the overall best performance for each task is highlighted in \textit{\textbf{ bold-italic}} font.     
\subsection{Sub-task A} 

Table \ref{tab:taskA} summarizes the obtained results. The results show that training the models on tweets as well as the non-sarcastic rephrases improve the performance for both languages, especially Arabic where an important performance increment is yielded.  Moreover, the performance of the evaluated models depends on the employed loss function. Although Model 1 and Model 2 are simple, they achieve better results than the GAN-based model (Model 3). The best performances on sub-task A are obtained using Model 2 in conjunction with the FL loss (0.6217) and Model 1 in conjunction with BCE loss (0.3833) for Arabic and English respectively. In our official submission, ensembling the models that are trained with and without the rephrase data harms the results. Our submitted system achieves the best performance on Arabic and ranked 15th on the English.

\subsection{Sub-task B} 
Table \ref{tab:taskb} presents our obtained results for sub-task B. Since we use the sarcastic tweets only for training, we only train the Model 3. The results show that the best performance is obtained using the BCE loss.  The FL and W. BCE loss functions have not improved the results. This might be explained by the fact that we did not tune the hyper-parameters $\alpha$ and $gamma$  of FL loss. Besides, in the W. BCE, the positive classes are assigned larger importance weights in comparison to the negative ones (see Section \ref{sec:obj}). Our official submission yields the second-best results on this sub-task.  

\subsection{Sub-task C}
Table \ref{tab:taskc} summarizes the obtained results on sub-task C test sets of Arabic and English languages. In accordance with the results of sub-task A, the models that are trained on the tweets and the non-sarcastic rephrases yield better performances.  The best-obtained accuracy scores are 0.85 and 0.71 in comparison to 0.78 and 0.69 obtained by our official submission for Arabic and English respectively. Hence, ensembling the models that are trained with and without the rephrase data harms the performance of our official submission. Our official submission is ranked 7th and 11th on Arabic and English respectively. 

\section{Conclusion}

In this paper, we present our participating system in the iSarcasmEval shared task for intended sarcasm detection in English and Arabic. Our system relies on three deep learning-based models that leverage two existing pre-trained language models for Arabic and English. We  participate in all sub-tasks, investigate several training objectives, and we study the impact of including non-sarcastic rephrase in the training data. The results show that ensembling models that are trained with and without rephrases have a negative impact on the official results. Our official submissions achieve the best performance on sub-task A for the Arabic language and rank in the second position on sub-task B.  

\section*{Acknowledgments}
Experiments presented in this paper were carried out using the supercomputer Toubkal, supported by Mohammed VI Polytechnic University (\url{https://www.um6p.ma}), and facilities of simlab-cluster HPC \& IA platform.
\bibliography{custom}

\end{document}